\documentclass[conference]{IEEEtran}
\IEEEoverridecommandlockouts
\usepackage{cite}
\usepackage{amsmath,amssymb,amsfonts}
\usepackage{algorithmic}
\usepackage{graphicx}
\usepackage{textcomp}
\usepackage{hyperref}
\usepackage{xcolor}
\def\BibTeX{{\rm B\kern-.05em{\sc i\kern-.025em b}\kern-.08em
    T\kern-.1667em\lower.7ex\hbox{E}\kern-.125emX}}

\usepackage{soul}
\usepackage{hyperref}
\usepackage{makecell}
\usepackage{multirow}
\usepackage{array}
\usepackage{caption}
\usepackage{subcaption}
\usepackage{placeins}
\usepackage{tabularx}
\usepackage{booktabs}
\usepackage{tablefootnote}
\usepackage{color}

\newcolumntype{P}[1]{>{\centering\arraybackslash}p{#1}}
\newcolumntype{M}[1]{>{\centering\arraybackslash}m{#1}}

\begin{document}

\title{IEEE BigData 2023\\ Keystroke Verification Challenge (KVC)\\

\thanks{This project has received funding from the European Union’s Horizon 2020
research and innovation programme under the Marie Skłodowska-Curie grant
agreement No. 860315, by project INTER-ACTION (PID2021-126521OB-I00 MICINN/FEDER), and by project HumanCAIC (TED2021-131787BI00 MICINN).}
}

\author{\IEEEauthorblockN{Giuseppe Stragapede\IEEEauthorrefmark{1}\thanks{Email: \href{mailto:giuseppe.stragapede@uam.es}{giuseppe.stragapede@uam.es}}, 
Ruben Vera-Rodriguez\IEEEauthorrefmark{1},
Ruben Tolosana\IEEEauthorrefmark{1},
Aythami Morales\IEEEauthorrefmark{1},
Ivan DeAndres-Tame\IEEEauthorrefmark{1},\\
Naser Damer\IEEEauthorrefmark{2},
Julian Fierrez\IEEEauthorrefmark{1},
Javier-Ortega Garcia\IEEEauthorrefmark{1},
Nahuel Gonzalez\IEEEauthorrefmark{3},
Andrei Shadrikov\IEEEauthorrefmark{4},
Dmitrii Gordin\IEEEauthorrefmark{5},\\
Leon Schmitt\IEEEauthorrefmark{6},
Daniel Wimmer\IEEEauthorrefmark{6},
Christoph Großmann\IEEEauthorrefmark{6},
Joerdis Krieger\IEEEauthorrefmark{6}, 
Florian Heinz\IEEEauthorrefmark{6},
Ron Krestel\IEEEauthorrefmark{6},\\
Christoffer Mayer\IEEEauthorrefmark{6},
Simon Haberl\IEEEauthorrefmark{6},
Helena Gschrey\IEEEauthorrefmark{6},
Yosuke Yamagishi\IEEEauthorrefmark{7},
Sanjay Saha\IEEEauthorrefmark{8}, \\
Sanka Rasnayaka\IEEEauthorrefmark{8},
Sandareka Wickramanayake\IEEEauthorrefmark{9}, 
Terence Sim\IEEEauthorrefmark{8},
Weronika Gutfeter\IEEEauthorrefmark{10},
Adam Baran\IEEEauthorrefmark{10},\\
Mateusz Krzyszto\'n\IEEEauthorrefmark{10} and
Przemysław Jask\'oła\IEEEauthorrefmark{10}
}


\IEEEauthorblockA{\IEEEauthorrefmark{1}Biometrics and Data Pattern Analytics (BiDA) Lab, Universidad Autonoma de Madrid, Spain}
\IEEEauthorblockA{\IEEEauthorrefmark{2}Fraunhofer Institute for Computer Graphics Research IGD, Darmstadt, Germany}
\IEEEauthorblockA{\IEEEauthorrefmark{3}Laboratorio de Sistemas de Información Avanzados (LSIA), Universidad de Buenos Aires, Argentina}
\IEEEauthorblockA{\IEEEauthorrefmark{4}Verigram LLC, Singapore}
\IEEEauthorblockA{\IEEEauthorrefmark{5}Citix, Almaty, Kazakhstan}
\IEEEauthorblockA{\IEEEauthorrefmark{6}Regensburg University Of Applied Sciences, Germany}
\IEEEauthorblockA{\IEEEauthorrefmark{7}Division of Radiology and Biomedical Engineering, The University of Tokyo, Japan}
\IEEEauthorblockA{\IEEEauthorrefmark{8}National University of Singapore, Singapore}
\IEEEauthorblockA{\IEEEauthorrefmark{9}University of Moratuwa, Sri Lanka}
\IEEEauthorblockA{\IEEEauthorrefmark{10}NASK – National Research Institute, Warsaw, Poland}

}


\maketitle

\IEEEpubidadjcol

\begin{abstract}
This paper describes the results of the IEEE BigData 2023 Keystroke Verification Challenge\footnote{Website: \href{https://sites.google.com/view/bida-kvc/}{https://sites.google.com/view/bida-kvc/}} (KVC), that considers the biometric verification performance of Keystroke Dynamics (KD), captured as \textit{tweet}-long sequences of variable transcript text from over 185,000 subjects. The data are obtained from two of the largest public databases of KD up to date, the Aalto Desktop and Mobile Keystroke Databases, guaranteeing a minimum amount of data per subject, age and gender annotations, absence of corrupted data, and avoiding excessively unbalanced subject distributions with respect to the considered demographic attributes.
Several neural architectures were proposed by the participants, leading to global Equal Error Rates (EERs) as low as 3.33\% and 3.61\% achieved by the best team respectively in the desktop and mobile scenario, outperforming the current state of the art biometric verification performance for KD. Hosted on CodaLab\footnote{\href{https://codalab.lisn.upsaclay.fr/competitions/14063/}{https://codalab.lisn.upsaclay.fr/competitions/14063/}}, the KVC will be made ongoing to represent a useful tool for the research community to compare different approaches under the same experimental conditions and to deepen the knowledge of the field.

\end{abstract}

\begin{IEEEkeywords}
Keystroke dynamics, behavioral biometrics, biometric verification, KVC, challenge
\end{IEEEkeywords}

\section{Introduction}
\label{sec:intro}
Keystroke Dynamics (KD) refers to the typing behavior exhibited by individuals and is commonly categorized as a \textit{behavioral biometric} trait, akin to voice \cite{voice}, signature \cite{TOLOSANA2022108609, 9335993}, gait \cite{PAMI_RVera2018, gait1, gait2}, touch gestures \cite{stragapede2022ijcb, stragapede2023behavepassdb}, and others. The implementation of keystroke dynamics verification systems is cost-effective, as it requires no additional hardware beyond what is already present in personal computers, laptops, smartphones, or tablets. Potential applications range from authenticating a user's identity while composing an email or taking an online test in free-text format \cite{hernandez2019edbb, rasnayaka2018wants}, to identifying malicious users across multiple accounts based on their typing style in free-text format \cite{typenet}, or serving as an additional biometric security layer alongside traditional knowledge-based passwords in fixed-text format \cite{kboc}, among other uses.

A broad classification of Keystroke Dynamics (KD) can be carried out based on two criteria: \textit{(i)} the type of acquisition device (keyboard), categorized as desktop or mobile, often kept distinct due to variations in the posture or typing activity of individuals; \textit{(ii)} with respect to the text format, which can be free, fixed, or transcript. In the first scenario, the typed text differs across various samples, resulting in sparser, more unstructured data with a higher incidence of typing errors, as opposed to the fixed-text scenario that seeks to simulate, for example, an intruder entering the victim's password. Lastly, the transcript text (considered in this study) can be characterized as a hybrid format, wherein subjects are instructed to read, memorize, and type a provided text.

In this scenario, we propose a novel experimental framework to benchmark KD for biometric verification in the form of the Keystroke Verification Challenge (KVC), hosted on CodaLab. Upon the submission of the results, the CodaLab platform returns several metrics (Sec. \ref{sec:metrics}) that quantify the recognition performance of biometric systems. The KVC, which will be made ongoing, is structured into two different tasks (desktop and mobile). To simplify the participation, a development and an evaluation dataset with a list of comparisons per task are provided to the participants, together with two Python scripts for loading the datasets, launching the training, and generating the scores to be submitted on CodaLab for evaluation.

\section{The Data}
\label{sec:the_data}

The proposed experimental framework relies on the two most comprehensive and extensive public databases of transcript-text keystroke dynamics available to date, collected by the User Interfaces\footnote{\href{https://userinterfaces.aalto.fi/}{https://userinterfaces.aalto.fi/}} group at Aalto University in Finland. These databases are gathered in a desktop\footnote{\href{https://userinterfaces.aalto.fi/136Mkeystrokes/}{https://userinterfaces.aalto.fi/136Mkeystrokes/}} \cite{Dhakal2018} and mobile\footnote{\href{https://userinterfaces.aalto.fi/typing37k/}{https://userinterfaces.aalto.fi/typing37k/}} \cite{palin2019people} acquisition environment, encompassing approximately 168,000 and 60,000 subjects, effectively capturing the challenges associated with widespread application usage. Each acquisition session involves a transcript text sentence (with variable content, though not entirely free-text). The data are captured through a web application in an unsupervised manner, reflecting real-world scenarios. Participants are instructed to read, memorize, and type English sentences on their devices, randomly chosen from a set of 1,525 sentences. Subject metadata, such as age and gender, are self-reported during the data acquisition process.

\section{Challenge Set Up}
\label{sec:challenge_setup}
This section provides a general outline of the challenge setup. For a more in-depth exploration of the dataset characteristics and a benchmark of the proposed evaluation framework, please consult \cite{stragapede2023kvc}.

Starting from the raw databases, a data preprocessing step is undertaken to organize them into a suitable format for Keystroke Dynamics (KD) analysis.
The raw data obtained include the timestamp of the moment a key is pressed, the timestamp of the moment the key is released, and the ASCII code of the key. Following the exclusion of certain subject data with insufficient acquisition sessions (fewer than 15 sessions per subject), the two downloaded databases are restructured to yield four datasets:
\begin{itemize}
    \item Desktop Dataset:
    \begin{enumerate}
        \item Development set: 115,120 subjects provided in a single .npy file that contains a Python nested dictionary (subject IDs: session IDs: data). Average session length: 48.65 ($\sigma$ = 18.50) characters typed.
        \item Evaluation set: data from 15,000 subjects, provided in a single .npy file that contains a shallow Python dictionary (sessions IDs: data). Average session length: 48.77 ($\sigma$ = 18.64) characters typed.
    \end{enumerate}
    \item Mobile Dataset:
    \begin{enumerate}
        \item Development set: 40,639 subjects provided in a single .npy file that contains a Python nested dictionary (subject IDs: session IDs: data). Average session length: 48.59 ($\sigma$ = 21.84) characters typed.
        \item Evaluation set: data from 5,000 subjects, provided in a single .npy file that contains a shallow Python dictionary (sessions IDs: data). Average session length: 47.98 ($\sigma$ = 20.93) characters typed.
    \end{enumerate}
\end{itemize}

The proposed experimental framework adheres to an open-set learning protocol, i.e., the subjects in the development and evaluation sets are distinct. Although an explicit validation set is not provided, it can be derived from the development set according to different training approaches. The demographic distribution of the datasets presented in the competition is detailed in \cite{stragapede2023kvc}.

The evaluation sets are separated by scenario (desktop and mobile), and they are provided in the form of two shallow Python dictionaries containing independent sessions. Such data are accompanied by the respective lists of pairwise comparisons to be carried out. Two Python script files are provided to load the data, and run the comparisons, generating a text file with the scores of each comparison, ready to be submitted to CodaLab for scoring.

\section{Experimental Protocol}
\label{sec:experimental_protocol}

The design and implementation of the evaluation protocol outlined in this section constitute a significant novelty introduced in this work.

Both tasks (desktop and mobile) follow a similar structure and are tailored for a biometric verification protocol. In essence, a score ranging from 0 to 1, indicative of a single comparison between two biometric samples, will be generated (1: same identity, 0: different identities). This poses a binary classification problem, as there is no need to determine the specific identity to which a given biometric sample belongs (identification). Within this experimental framework, a biometric sample corresponds to an acquisition session.

The total number of 1 vs 1 session-level comparisons is as follows:
\begin{itemize}
    \item Task 1 (Desktop): 2,250,000 comparisons, involving 15,000 subjects not included in the development set.
    \item Task 2 (Mobile): 750,000 comparisons, involving 5,000 subjects not included in the development set.
\end{itemize}

For each subject, there are 5 enrolment sessions and 10 verification sessions, resulting in 50 1vs1 comparisons. These are averaged over the 5 enrolment sessions, producing 10 genuine scores per subject. Similarly, 20 impostor scores per subject are generated. The impostor sessions are categorized into two groups: 10 \textit{similar} impostor scores, where the verification sessions are randomly chosen from subjects within the same demographic group (same gender and age group); and 10 \textit{dissimilar} impostor scores, where the verification sessions are all randomly selected from subjects of different gender and age intervals.

Based on the described evaluation design, we consider two cases for evaluating the system:
\begin{itemize}
    \item Global distributions: this case corresponds to dividing all scores into two groups, genuine and impostor scores, regardless of which subject they belong too. This case corresponds to having a fixed, pre-determined threshold, implying a simpler deployment of the biometric system. In order to assess the performance of the biometric system, this choice means setting one single threshold for all comparisons to obtain a decision. 
    \item Mean per-subject distributions: the optimal threshold is computed at subject-level, considering the 30 verification scores as described above. This choice corresponds to
    providing the system with more flexibility, so that it can adapt to user-specific distributions \cite{Fierrez-Aguilar2005_ScoreNormalization, Fierrez-Aguilar2005_AdaptedMultimodal}. In a real-life use case, this would require processing the subject's enrolment samples to establish a threshold. 
    It is important to highlight that this does not require re-training or fine-tuning the biometric system using subject-specific data. Then, all metrics computed per-subject are averaged considering all subjects in the evaluations set to obtain the values displayed. Generally, the verification performance of the system benefits from considering a different threshold per user. 
\end{itemize}

\section{The Metrics}
\label{sec:metrics}
Over the years, various metrics have been proposed for biometric verification. The commonality among these metrics is their reliance on (normalized) scores typically generated through pairwise comparisons of biometric samples for biometric verification. To ensure a fair comparison between systems, different metrics are computed within the challenge. These include the global Equal Error Rate (EER) (\%), mean per-subject EER (refer to Sec. \ref{sec:experimental_protocol}), global False Non-Match Rate (FNMR) at 1\% False Match Rate (FMR), FNMR at 10\% FMR, and AUC (Area Under the ROC Curve). Additionally, Detection Error Tradeoff (DET) curves are provided. A brief explanation of the various metrics is provided below.

A false match (FM) is defined as a comparison decision indicating a match for a biometric probe and a biometric reference that are from different biometric capture subjects, while a false non-match (FNM) is defined as a comparison decision indicating a non-match for a biometric probe and a biometric reference that are from the same biometric capture subject and of the same biometric characteristic. The rates associated with FMR (FNMR) correspond to the proportion of completed biometric non-mated (mated) comparison trials resulting in a false match (non-match) \cite{ISObiometrics}.

\subsubsection*{Equal Error Rate (EER)} The EER describes the point in which the FMR and FNMR curves intersect. The two rates typically have opposite trends with respect to the threshold setting (in the case of genuine scores closer to 1, and impostor scores closer to 0, as the threshold of a biometric system increases, the FMR will drop and the FNMR curve will rise). 

\subsubsection*{False Non-Match Rate at X\% False Match Rate (FNMR @ X\% FMR)} We consider $X = 1\%, 10\%$. This also corresponds to a point on the DET curve. This metric expresses a trade-off between security and usability \cite{ISO1998}. In fact, while from the point of view of security the priority is avoiding intrusions, denying the access to the genuine subject a large number of times would generate frustration and highly impacts the usability of the system. In this case, the threshold is set to X = 1\%, 10\% of FMR (rejection of 99\%, 90\% of impostor attempts, respectively), aiming to minimize the FNMR.

\subsubsection*{Area Under the Receiver Operating Characteristic (ROC) Curve (AUC)} The ROC curve is the plot of the TMR (True Match Rate) against FMR, at various threshold settings. A true match corresponds to the case of a genuine subject recognized as such. By definition, the TMR and the FNMR sum to 1. A perfect classifier has an Area Under the ROC Curve (AUC) of 1. 

\subsubsection*{Detection Error Trade-off (DET) curve} The DET curve plots FMR against FNMR, at various threshold settings, typically on a non-linear scale. As the threshold decreases, the amount of false matches (impostor subjects classified as genuine) increases, and the number of false non-matches decreases (genuine subjects classified as impostor). The closest the DET curve to the bottom left corner, the better the biometric system will be. The intersection of a DET curve with the line $y = x$ corresponds to the EER.

\section{The Biometric Systems Proposed}
\label{sec:biometric_systems}
A summary of the biometric systems participating in the challenge and the main results are shown in Table \ref{tab:archi_comp}. The challenge ranking is created and displayed in the Table for each task based on the global EER (\%). The LSIA team is the winner of both tasks. In the remainder of this section, the biometric verification systems proposed by all teams are presented. Results according to different metrics are presented and discussed in Sec. \ref{sec:experimental_results}.

\begin{table*}[h!]
\begin{minipage}{\linewidth}
\centering
\scriptsize
\caption{\small High-level comparison of the proposed keystroke biometric verification systems. \textit{D} stands for desktop, \textit{M} stands for mobile.}
\begin{tabular}{M{0.11\textwidth}|M{0.25\textwidth}|M{0.16\textwidth}|M{0.05\textwidth}|M{0.05\textwidth}|M{0.05\textwidth}|M{0.05\textwidth}}
 \cmidrule[.75pt]{4-7}
\multicolumn{3}{c}{} & \multicolumn{2}{c|}{\textbf{Desktop}} & \multicolumn{2}{c}{\textbf{Mobile}}  \\ 
\bottomrule
\makecell[c]{\textbf{System}} & \makecell[c]{\textbf{Architecture}} & \makecell[c]{\textbf{Loss Function}} & \makecell[c]{\textbf{Global}\\ \textbf{EER (\%)}} & \textbf{Position} & \makecell[c]{\textbf{Global}\\ \textbf{EER (\%)}} & \makecell[c]{\textbf{Position}} \\
\toprule
LSIA & CNN+RNN with attention mechanism & Custom & 3.33 & 1 & 3.61 & 1\\
VeriKVC & CNN & ArcFace & 4.03 & 2 & 3.78 & 2 \\
Keystroke Wizards & GRU (\textit{D}), Transformer (\textit{M}) & Triplet & 5.22 & 3 & 5.83 & 5 \\
U-CRISPER & GRU-Based Siamese Network & Triplet & 6.19 & 4 & 8.76 & 6 \\
YYama & Transformer+CNN & Contrastive + Cross-entropy & 6.41 & 5 & 4.16 & 3 \\
Challenger$\ast$\footnotemark & Transformer & Triplet & 6.79 & 6 & 5.19 & 4 \\
BioSense & CNN with attention mechanism & Cross-entropy & 10.85 & 7 & 11.83 & 7 \\
\bottomrule
\end{tabular}
\label{tab:archi_comp}
\end{minipage}
\vspace{0.2cm}

\end{table*}

\subsection{Team: LSIA}

The LSIA team is composed of one member from Laboratorio de Sistemas de Informaci\'on Avanzados (LSIA), Universidad de Buenos Aires (Argentina).

\subsubsection{Model architecture}

Among the architectures proposed by the LSIA team, the best performance in terms of EER is achieved by a dual-branch (recurrent and convolutional) embedding model for distance metric learning. The same architecture is adopted for both tasks. 

The recurrent branch comprises two bidirectional GRU layers (512 units), while the convolutional branch features three blocks of 1D convolution, each with an increasing number of filters (256, 512, and 1024, with kernel size equal to 6) and utilizes global average pooling. Temporal attention serves as the first layer of both branches. Scaled dot-product attention is applied between the recurrent layers, and channel attention follows each convolutional layer. The outputs of both branches are concatenated, and the final embedding vector is produced by three dense layers. Batch normalization and dropout are employed after each processing layer.

Sample similarity is measured, as is customary in distance metric learning, using the Euclidean distance between their respective embeddings. 

The choice of architecture is motivated by the observation that keystroke timings result from a combination of two factors: a partially conscious decision process involving what to type and an entirely unconscious motor process pertaining to how to type \cite{gonzalez2021shape}. The convolutional branch is expected to excel at identifying common, short sequences, while the recurrent branch is expected to capture the user’s time-dependent decision process. Empirical testing confirms that a dual-branch model outperforms a purely recurrent or purely convolutional model with an equivalent number of trainable parameters.

\subsubsection{Training}

For training the model, only the development sets provided by the organizers are used. These sets are preprocessed to generate five attributes: the integer key code (ASCII), for which the network learns a small-dimensional embedding, and four normalized timing features, using a scale of seconds. These are the interval between key press and release events, called the Hold Time (HT), the latency between successive key press events, called Inter-Press Time (IPT), also known as Flight Time (FT), and two synthetic features (SHT and SFT) meant to capture variations in the user's typing style compared to the general population \cite{gonzalez2022towards}. 

A novel loss function is used to minimize the EER of a keystroke dynamics verification system using one-shot evaluation with a uniform global detection threshold across all users. This function extends the SetMargin loss of Morales \textit{et al.} \cite{2022_PR_SetMargin_Morales} to sets of sets (rather than pairs of sets) and includes an additive penalty term to encourage the model to embed all classes within hyperspheres with similar average radii. 

A learning curriculum of increasing difficulty is adopted to train the model. Each batch consisted of 15 samples from 40 different users, totaling 600 samples per batch. In each batch of epoch \textit{i}, the first user is paired with its \textit{i} nearest neighbors based on the proximity of its embedding centers.

The tools and frameworks used to train the model included Keras 2.6.0 and TensorFlow 2.12.0, running in Python 3.10.10, and making use of an NVIDIA A100 80GB GPU. The synthetic timing features are generated using the KSDSLD tool, which is publicly available at \cite{gonzalez2023ksdsld}.

\subsection{Team: VeriKVC}

The VeriKVC team is composed of two members from two different companies, respectively Verigram LLC (Singapore), and Citix (Kazakhstan).

\subsubsection{Model Architecture}
The system architecture proposed by the VeriKVC team relies on a Convolutional Neural Network (CNN) design, due to its capability to effectively capture spatial features in web behavior data. The CNN model is tailored to process session data and extract significant patterns and features. The same architecture is adopted for both tasks. 

\footnotetext{Having used the raw data of the Aalto Databases, from which the KVC datasets are obtained, it is likely that some of the subjects in the evaluation set were also used for the pre-training, lifting the open set learning protocol restriction, according to which the development and evaluation sets should not have any subject in common. A dedicated scenario, called Unrestricted, is included in the KVC accounting for the option of pre-trained models. Being this the only team that participated in this form, it is included in the general rankings, but with a special mark $\ast$.}

In addition to the chosen architecture, feature engineering techniques are employed to preprocess the behavior data. One such technique involves encoding the timing differences of events using both sine and cosine functions \cite{vaswani2017attention}. This normalization helps to account for the varying time scales of events and maintain the relevance of timing information. Furthermore, the difference between the end and start times of events is processed using cosine functions to scale it within the range of -1 to 1. These features extracted by the attention mechanism are essential in providing the model with a better understanding of the temporal aspects of web behavior data.

In response to the potential risk of overfitting, various randomization techniques are utilized for a data augmentation process during training. These techniques introduce variability into the training data, preventing the model from memorizing the training set and promoting its generalization to unseen data. By enhancing the model robustness through data augmentation, its ability to handle diverse real-world scenarios effectively is ensured.

\subsubsection{Training}

To train the model, the number of epochs was set to 5000 with a batch size of 512. This extended training duration allows the model to learn complex patterns and improve its overall performance. The AdamW optimizer\cite{loshchilov2017decoupled} with learning rate of 0.01 and gradient clipping and cosine annealing scheduler \cite{loshchilov2016sgdr} is chosen. This allows the network to learn basic features quickly and spend more time on tuning more subtle features. In terms of loss function, ArcFace \cite{deng2019arcface} is selected as the most effective choice for the model. This loss encourages the model to learn distinct feature representations for each individual, which is crucial for subject verification tasks. 

The VeriKVC team believes in the importance of open collaboration and knowledge sharing. The code of the described solution is available online on the GitLab page\footnote{\href{https://gitlab.com/vuvko/kvc}{https://gitlab.com/vuvko/kvc}}, where it is possible to review the proposed implementation, facilitating transparency and enabling others to replicate and build upon their work.

\subsection{Team: Keystroke Wizards}

The Keystroke Wizards team is composed of members of the Regensburg University of Applied Sciences (Germany). This is one of the two teams from the same institution\footnote{Team members: Leon Schmitt, Daniel Wimmer, Christoph Großmann, Joerdis Krieger, and Florian Heinz.}. 

\subsubsection{Model Architecture}

In the desktop task, the proposed model is based on a Recurrent Neural Network (RNN) combined with a triplet loss function, following the approach in \cite{typenet}. In contrast, in the mobile task, the proposed model is based upon the models proposed in \cite{typeformer}, \cite{FG}, and \cite{senerath2023behaveformer}, which all use Gaussian Range Encoding (GRE) as well as triplet loss.

The desktop model employs a Multi-Layer Gated Recurrent Unit (GRU) RNN with 11 input features: HT, HT of the following key, IPT, time from first key press to following key release, Inter-Key Time (IKT), Inter-Release Time (IRT), rollover duration, rollover count, hold-to-rollover ratio, ASCII code, and ASCII code of following key. The rollover-duration feature describes for how long both keys of the key pair are pressed concurrently. The rollover-count describes how often both keys are pressed at the same time. The hold-to-rollover-ratio describes the ratio of the total hold times to the total rollover times of the session. These features are chosen based on data analysis, correlation analysis, and scatter plots.

The mobile model is a hybrid Transformer with a Channel Module and a Temporal Module, incorporating Multi-Scale CNN and GRU layers. In addition to desktop features, they use trigram features, capturing the timing of the first and third key presses. The mobile model integrates GRUs for temporal data and CNNs for spatial data. It features multi-head attention and Gaussian range positional encoding. 

\subsubsection{Training}

Both models are trained with the Triplet Loss function. The desktop model utilizes a PyTorch-based architecture featuring batch normalization, GRU layers, dropout, and linear layers. Training parameters include 160 epochs, 128 batches per epoch, 1024 sequences per batch, a sequence length of 70, and 80\% data for training. The mobile model is trained with 160 epochs, a sequence length of 50, and 80\% training data.

The hardware used for model training and evaluation consists of a GeForce RTX 3090 graphic card, an AMD Ryzen 9 5950X 16-core processor, and 64 GB of RAM.

\subsection{Team: U-CRISPER}
The U-CRISPER team is composed of members from the Regensburg University of Applied Sciences (Germany). This is one of the two teams from the same institutions\footnote{Team members: Ron Krestel, Christoff Mayer, Simon Haberl, Helena Gschrey}. 

\subsubsection{Model Architecture}

Four essential time-based features are derived: HT, IKT, IPT, and IRT. These features are standardized by removing the mean and scaling to unit variance. Any missing value is substituted by the mean. The ASCII codes, normalized by dividing them by 255, are also used. The raw data also contain upward and downward outliers, which could potentially introduce confusion into the model. To address this issue, an in-depth analysis of the data point distribution is conducted. As a result, extreme outliers are clipped to a predefined boundary, enhancing the model's generalization.
The adopted model architecture is a GRU-Based Siamese Network, utilizing Gated Recurrent Units (GRU) to adaptively capture temporal dependencies \cite{chung2014empirical}. Therefore, the design fits well for handling keystroke sequences. Two GRU layers, each with a hidden size of 64 and a dropout rate of 0.2 in between, process an input sequence (70 by 5) and a Linear Layer with a prior Rectified Linear Unit (ReLU) activation generates an embedding vector of dimension 64.

\subsubsection{Training}

For effective training, the Triplet Loss function \cite{typenet} is used. Each comparison is carried out at session level: one session is used as an anchor, one as a positive from the same user, and one as a negative from a different user. The model processes each session, generating embeddings compared by the Triplet Loss function, which minimizes the anchor-positive distance and maximizes the anchor-negative distance.

An hyperparameter optimization is conducted by using Optuna's Tree-structured Parzen Estimator (TPE) \cite{watanabe2023tree}, resulting in the following model configuration: 200 epochs, 70-length input sequences,  batch size of 512 (150 for training, 40 for validation), Adam optimizer (LR: 0.001, betas: 0.9, 0.999, epsilon: 1e-08), and Triplet Loss (margin: 1.5, p-value: 2.0).

\subsection{Team: YYama}
The YYama team is composed of one member from the Division of Radiology and Biomedical Engineering, The University of Tokyo (Japan).

\subsubsection{Model Architecture}

The proposed ``dual-network'' approach combines an embedding model for feature extraction and a classifier model for verification, aiming to improve the biometric verification performance.

The development dataset is randomly split into 80\% for training and 20\% for validation. The feature extraction stage included five conventional variables (HT, IKT, IPT, IRT, and ASCII code) \cite{typenet}, \cite{typeformer}, plus two new variables (differential in consecutive HTs and ASCII codes). These features are standardized and normalized to maintain consistency across the dataset.

The same architecture is used for both tasks. The embedding model architecture is a simplified transformer-based design featuring 1D CNN layers followed by a transformer encoder and a max pooling operation, concluding with a fully connected layer. In contrast, the classifier model is a straightforward 1D CNN-based architecture that took paired outputs from the embedding model to determine if they originate from the same user.

\subsubsection{Training}
The training settings are consistent across both tasks, with the embedding model using the Adam optimizer and a fixed learning rate of 0.001, trained for 2,000 epochs for mobile and 1,000 epochs for desktop. The batch size during training is 2,048, with a total of 16 batches supplied to the models per epoch. Therefore, the total number of data pairs per epoch was 32,768. The loss function is contrastive loss with a margin of 1 for the embedding model. Simultaneously, the classifier model's embeddings are trained using the binary cross-entropy loss function with the same learning rate as the embedding model.

Model performance is evaluated using the global EER, and the best models for mobile and desktop are selected based on the lowest EER. The dual-network approach yields lower EERs than conventional approaches based on Euclidean distance: 5.65 vs. 13.07 locally and 6.41 vs. 14.35 on the leaderboard for desktop tasks, and 3.09 vs. 7.47 locally and 4.16 vs. 9.75 on the leaderboard for mobile tasks.

The hardware used was equipped with an Intel Core i9-10900k processor clocked at 3.70GHz and an NVIDIA RTX 3090 graphics card with 24GB of VRAM.

The experimental environment was set up on Ubuntu 22.04 LTS with Python 3.10.11. The libraries used included PyTorch 2.0.1, OpenCV 4.8.0.76, NumPy 1.24.3, and scikit-learn 1.3.0.

\begin{table*}[!h]
\scriptsize
\caption{\small Comparison of the results achieved in the Desktop and Mobile tasks considered in KVC.}
\begin{minipage}{\textwidth}
\centering
\begin{tabular}{P{0.05\textwidth}|P{0.11\textwidth}|P{0.11\textwidth}|P{0.1\textwidth}|P{0.1\textwidth}|P{0.1\textwidth}|P{0.1\textwidth}}
\multicolumn{7}{c}{\textbf{Desktop}} \\
\toprule
\textbf{Position} & 
\textbf{Team} & 
\makecell{\textbf{Global}\\ \textbf{EER$\downarrow$ (\%)}} &
\makecell{\textbf{FNMR$\downarrow$ @1\%}\\ \textbf{FMR (\%)}} & 
\makecell{\textbf{FNMR$\downarrow$ @10\%}\\ \textbf{FMR (\%)}} &
\textbf{AUC$\uparrow$ (\%)} &
\makecell{\textbf{Mean}\\ \textbf{Per-Subject}\\ \textbf{EER$\downarrow$ (\%)}} \\
\bottomrule
1 & LSIA & 3.33 & 11.96 & 0.51 & 99.48 & 0.77 \\
2 & VeriKVC & 4.03  & 18.79 & 1.05 & 99.07 & 1.32 \\
3 & Keystroke Wizards & 5.22 & 27.98 & 1.62 & 98.79 & 1.78 \\
4 & U-CRISPER & 6.19 & 35.24 & 2.68 & 98.37 & 2.44 \\
5 & YYama & 6.41 & 36.96 & 2.88 & 98.28  & 2.54\\
6 & Challenger$\ast$\footnotemark[11] & 6.79 & 39.36 & 3.52 & 98.09  & 2.8 \\
7 & BioSense & 10.85 & 54.59 & 12.0 & 95.86  & 5.17\\
\bottomrule
\end{tabular}

\vspace{0.1cm}
\begin{tabular}{P{0.05\textwidth}|P{0.11\textwidth}|P{0.11\textwidth}|P{0.1\textwidth}|P{0.1\textwidth}|P{0.1\textwidth}|P{0.1\textwidth}}
\multicolumn{7}{c}{\textbf{Mobile}} \\
\bottomrule
\textbf{Position} & 
\textbf{Team} & 
\makecell{\textbf{Global}\\ \textbf{EER$\downarrow$ (\%)}} &
\makecell{\textbf{FNMR$\downarrow$ @1\%}\\ \textbf{FMR (\%)}} & 
\makecell{\textbf{FNMR$\downarrow$ @10\%}\\ \textbf{FMR (\%)}} &
\textbf{AUC$\uparrow$ (\%)} &
\makecell{\textbf{Mean}\\ \textbf{Per-Subject}\\ \textbf{EER$\downarrow$ (\%)}} \\
\bottomrule
1 & LSIA & 3.61 & 17.44 & 0.6 & 99.28  & 1.03\\
2 & VeriKVC & 3.78 & 18.39 & 0.95 & 99.04 & 1.35\\
3 & YYama & 4.16 & 24.41 & 0.72 & 99.09 & 1.66 \\
4 & Challenger$\ast$\footnotemark[11] & 5.19 & 32.89 & 1.55 & 98.69 & 2.17 \\
5 & Keystroke Wizards & 5.83 & 41.58 & 1.93 & 98.34 & 2.66 \\
6 & U-CRISPER & 8.76& 67.15 & 6.68 & 96.54  & 5.07 \\
7 & BioSense & 11.83 & 60.48 & 14.43 & 94.83  & 5.75\\
\bottomrule
\end{tabular}
\end{minipage}
\vspace{0.2cm}
\label{tab:all_metrics}
\end{table*}

\subsection{Team: Challenger}
The Challenger team is composed of three members from the National University of Singapore, and one member from the University of Moratuwa (Sri Lanka).

\subsubsection{Model Architecture}
Starting from the raw data, some features are extracted considering digram and trigram features. The time intervals between various events to create features are also used. The combination of these time intervals and ASCII codes yields a total of 10 keystroke features. ASCII code is normalized to the range [0,1] and time based features are represented in seconds.

While deep learning techniques like RNNs and LSTMs excel in behavioral biometrics, Transformer architectures have gained prominence. However, applying Transformers to keystroke data faces challenges due to limited data and multimodality. To address this, a Spatio-Temporal Dual Attention \cite{senerath2023behaveformer} method for improved attention and feature extraction across time and modalities is employed. The proposed model is based on a variation of the Vanilla Transformer encoder named Spatio-Temporal Dual Attention Transformer, that is based on a framework named BehaveFormer \cite{vaswani2017attention, senerath2023behaveformer}. In this framework, the raw data are preprocessed to extract additional features. Then, the Transformer architecture with Dual Attention modules are used to extract more discriminative features from KD. Then, the Spatio-Temporal Dual Attention Transformer uses a single transformer with two attention mechanisms: one over the temporal axis and one over the channel axis. This allows the model to focus on the relevant keystroke features over time, extracting unique behavioral patterns for individual users. The GRE is used as the positional encoding. Moreover, it uses a 2D convolution network instead of the feed forward network in the transformer blocks, to analyze the input data over time and across different channels.


\subsubsection{Training}

To give the model a better starting point than the randomly assigned weights, an initial model is pre-trained which is a good generalized starting point for the keystroke verification task, providing a solid base for the further training on the data provided by the organizers. For the pre-training, keystroke data from these datasets are used: Aalto Databases \cite{Dhakal2018, palin2019people}, HMOG \cite{yang2014multimodal}, and HuMidb \cite{acien2021becaptcha}. 
The Triplet Loss function is used for training. Two different models with the same architecture are trained for the two tasks. The models were trained with a batch size of 512 and for 1050 epochs for the desktop model and 2500 epochs for the mobile .

\subsection{Team BioSense}
The BioSense team is composed of members from the NASK – National Research Institute (Warsaw, Poland).

\subsubsection{Model Architecture}

A model of a neural network with Keycode Attention, named Anabel-KA, is developed. Keycode Attention refers to the self-attention \cite{vaswani2017attention} module, in which the attention matrix is computed from the sequence of ASCII codes for the keyboard buttons pressed by the users, and values in the attention function are derived from the time intervals between pressing and releasing keys in the sequence. $Q$ and $K$ (in the equation of scaled dot product attention below) are a result of the keycode extraction function $F_{KB}$ and $V$ is a result of the time-interval extraction function $F_{T}$.

\begin{align*}
\label{eq:attention}
Att(Q,K,V)=&softmax(\frac{QK^T}{\sqrt{d_k}})V\\ 
&\mbox{where }Q=K=F_{KB}(X),V=F_{T}(X) \notag
\end{align*}

Feature extraction $F_{KB}$ from the keycodes is made by a set of 2D convolutions with kernels of size 3, as authors of other methods described in the literature~\cite{userauth},\cite{gonzalez} show the feasibility of analyzing 3-element groups of keys for the keystroke biometrics. Typing dynamics in the proposed model are represented by the sequence of HTs and IKTs. Such values are processed by the time extractor module $F_{T}$ which is also built using 2D convolutions of size 3 and stride 1. In both cases, the time extractor and the keycode extractor, the number of convolution filters is chosen to be 64. Input sequences are normalized and the length is arbitrarily set to 66, as the median length of the inputs in the KVC database is equal to 48 for mobile and desktop entries. Longer sequences are cut and shorter ones are padded with zeros. Extension of the model with the method for handling longer sequences is planned to be added in the future. Embeddings generated by the Anabel-KA have a length of 256. The Keycode Attention is implemented with the multi-head attention module with 8 heads. 

\subsubsection{Training}
For training, an additional classification layer is used with a number of outputs equal to the number of classes and softmax function applied. When evaluating, the similarity between keystroke sequences is computed using the Euclidean distance between their embeddings. Training is done with the cross-entropy loss function. For the purpose of the experiments, a cross-validation with 10 folds is employed, with 500 subjects excluded from the dataset for validation in each fold. The maximal number of epochs is set to 60, the final model is selected based on the maximal value of the validation metric, which is the AUC. The batch size is 128 and optimization is done with the Adam algorithm, the training procedure starts from the warm-up learning rate of 0.01 and changes gradually after every 10 epochs. The code of the Anabel-KA is available online\footnote{\href{https://github.com/nask-biometrics/anabel-ka}{https://github.com/nask-biometrics/anabel-ka}}.

\footnotetext[11]{Please check footnote \footnotemark[6].}

\section{Experimental Results}
\label{sec:experimental_results}
\begin{figure*}[!t]
\centering
\begin{subfigure}{.45\textwidth}
  \centering
  \includegraphics[width=\linewidth]{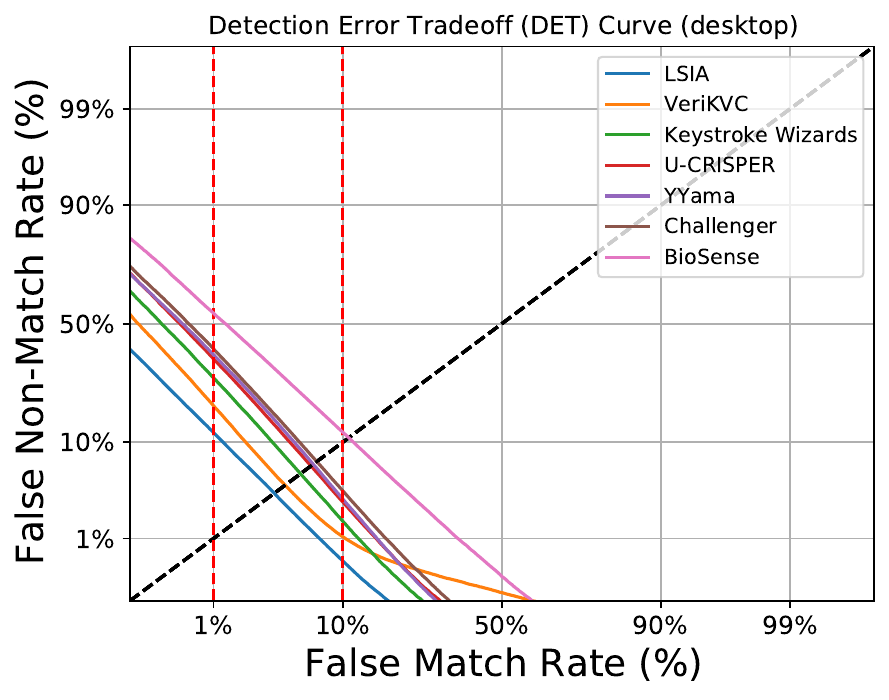}
  \caption{Desktop task.}
  \label{fig:det_desktop}
\end{subfigure}%
\begin{subfigure}{.45\textwidth}
  \centering
  \includegraphics[width=\linewidth]{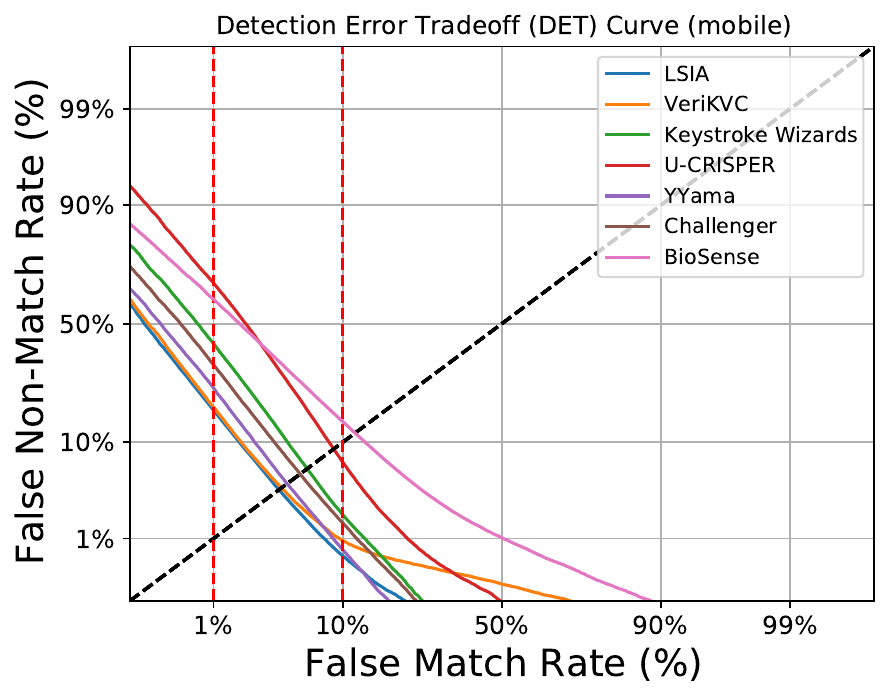}
  \caption{Mobile task.}
  \label{fig:det_mobile}
\end{subfigure}
\caption{DET curves including the results of all the biometric verification systems proposed in the KVC challenge for both desktop and mobile tasks. The red dashed lines indicate the operational points 1\% FMR and 10\% FMR whereas the black dashed line indicate the points where the FMR = FNMR.}
\label{fig:dets}
\end{figure*}

Table \ref{tab:all_metrics} presents the biometric verification results according to the different tasks and metrics.
Each row shows a different system, sorted by their ranking in the challenge.
As can be seen in the table, three teams achieve better results in the desktop task in comparison with the mobile task (LSIA, Keystroke Wizards, BioSense), while the remaining four (VeriKVC, U-CRISPER, YYama, Challenger) perform better in the mobile task. Consequently, the KVC results do not reflect the typical lower variability associated with the desktop task due to a more constrained acquisition scenario (i.e., in contrast to mobile devices, subjects are more likely to be sitting down and in a still position while typing on a desktop keyboard). Nevertheless, the best result is achieved in the desktop task with a global EER of 3.33\%.

The best performing team of both tasks, LSIA, achieves a global EER of 3.33\% and 3.61\% respectively in the desktop and mobile task, with a solid gap with respect to the second best team of both tasks, VeriKVC, especially in the desktop task (3.33\% vs. 4.03\%). It is interesting to point out that both teams use more sophisticated loss functions in comparison with other teams. In fact, the LSIA team adopts an extension of the SetMargin loss \cite{2022_PR_SetMargin_Morales} with an additive penalty term combined with a learning curriculum of increasing difficulty, whereas the VeriKVC team applies the Additive Angular Margin Loss (ArcFace) \cite{deng2019arcface}. Such approaches outperform current state-of-the-art performance on the same KVC experimental conditions, evidencing the importance of optimal loss functions for training \cite{stragapede2023kvc}. Moreover, the performance gap between these two teams might be given by the complexity of the model architecture adopted, i.e., the LSIA team uses a dual-branch (recurrent and convolutional) embedding model for distance metric learning, whereas the VeriKVC uses an attention module as feature extractor followed by a CNN. More in-depth analyses combining the two architectures could be carried out in the future to shed light on which are the most impactful aspects in learning discriminative features.

Another interesting aspect to point out is related to the fact that the ranking in both desktop and mobile tasks is consistent across the two evaluation scenarios considered, i.e., global EER and mean per-subject EER. The adoption of per-subject EER is generally associated with better results as different thresholds are considered per subject, allowing the system to better adapt to each subject. It is also important to remark that, in terms of the mean per-subject EER, the system developed by the LSIA team achieves results under 1\% in the desktop task (0.77\%), and very close in the mobile task (1.03\%).

The ranking is also consistent across the different metrics. For example, examining different threshold operating points (FNMR @1\% FMR and FNMR @10\% FMR), as well as the AUC, it is possible to see that the separation between the performance of different systems is clear. 

Concerning the other teams, the performance achieved is very satisfactory in both desktop and mobile tasks as well, with global EER results between 5.22\% (Keystroke Wizards) and 10.85\% (BioSense) for the desktop task and between 4.16\% (YYama) to 11.83\% (BioSense) for the mobile task. Considering the variety of the architectures proposed (RNNs, CNNs, Transformers), it is possible to conclude that biometric verification based on KD is a problem that can be tackled from several directions with good results. In particular, approaches based on distance metric learning (triplet and contrastive loss functions) used by Keystroke Wizards, U-CRISPER, YYama, and Challenger, seem to work better than cross-entropy loss (BioSense).

For completeness, we include in Figure \ref{fig:dets} a comparison of the biometric verification systems in terms of the Detection Error Tradeoff (DET) curves, for both desktop (Fig. \ref{fig:det_desktop}) and mobile (Fig. \ref{fig:det_mobile}) tasks. In the graphs, three dashed lines are marked: the black one represents the points where the FMR = FNMR (i.e., the EER metric). In contrast, the two red lines represent two operational points in which the errors at the intersection with the DET curves are unbalanced. Specifically, they respectively represent the score of FMR @1\% FNMR, and FMR @10\% FNMR (also reported in Table \ref{tab:all_metrics}). These tradeoffs are related to the threshold and the usability of the system: by setting a threshold that would let in only 1\% of the impostor attempts, the amount of genuine users who are denied the access is measured. The higher these values are, the less \textit{usable} the system would be, as false non-matches would generate frustration in the legitimate subjects.

Considering the DET curves, it is possible to observe that in the mobile task, three systems exhibit a change in the slope as the FMR increases (VeriKVC, U-CRISPER, BioSense), while a similar trend is visible in the desktop task only for VeriKVC. Having fixed the FNMR (considering a horizontal line on the DET curve graph), worse security of the system corresponds to higher values of FMR. Consequently, such trends highlight the deterioration of the biometric performance of a given system. Such trend is especially accentuated in the case of VeriKVC (based on a convolutional architecture), despite of being the second best system in terms of EER. 

In general terms, it can be observed that in comparison with recent literature, the results achieved in the KVC outperform the state of the art \cite{stragapede2023kvc}, proving the effectiveness of the discriminative power of transcript text KD as behavioral biometric characteristic. In fact, almost all teams achieve EER values lower than 10\% for both tasks. From this perspective, the most interesting metrics for future works could be represented by FNMR @1\% FMR as it can better separate the performance of the systems for high-security scenarios. Moreover, perhaps even considering the next operating point 0.1\% FMR, reducing the gap with other more popular physiological biometric traits such as face, for example.

\section{Conclusions and Future Work}
\label{sec:conclusions}
This paper presented the results of the Keystroke Verification Challenge (KVC). Within this limited-time challenge, current state of the art of KD-based biometric verification were outperformed, reaching EER values as low as 3.33\% and 3.61\% respectively in the desktop and mobile task by the winner team, LSIA.

The challenge is hosted on CodaLab\footnote{\href{https://codalab.lisn.upsaclay.fr/competitions/14063/}{https://codalab.lisn.upsaclay.fr/competitions/14063/}} and it is meant to be ongoing to represent a useful tool for the entire research community. In fact, a sound experimental framework was designed for the challenge, based on the two most complete and large-scale public databases of free-text keystroke dynamics up to date, collected respectively in a desktop and mobile acquisition environment, including keystroke data from more than 185,000 subjects overall. To this end, KVC aims to be a dedicated and unified test bench to foster the design of innovative solutions that achieve improved performance in comparison with existing ones.

Regarding future work, we plan to extend the considerations to other aspects of the developed systems, such as the biometric fairness of the systems, the number of parameters in the models, deploying the proposed models in Continuous Authentication (CA) schemes, and incorporating new evaluation sets to assess the generalization ability of the models across different acquisition scenarios and text formats. Furthermore, the exploration of approaches based on generating synthetic subject-specific data will be considered to evaluate their suitability for the problem of behavioral biometrics-based verification.

\bibliographystyle{IEEEtran}
\bibliography{0_Main}

\begin{thebibliography}{10}
\providecommand{\url}[1]{#1}
\csname url@samestyle\endcsname
\providecommand{\newblock}{\relax}
\providecommand{\bibinfo}[2]{#2}
\providecommand{\BIBentrySTDinterwordspacing}{\spaceskip=0pt\relax}
\providecommand{\BIBentryALTinterwordstretchfactor}{4}
\providecommand{\BIBentryALTinterwordspacing}{\spaceskip=\fontdimen2\font plus
\BIBentryALTinterwordstretchfactor\fontdimen3\font minus \fontdimen4\font\relax}
\providecommand{\BIBforeignlanguage}[2]{{%
\expandafter\ifx\csname l@#1\endcsname\relax
\typeout{** WARNING: IEEEtran.bst: No hyphenation pattern has been}%
\typeout{** loaded for the language `#1'. Using the pattern for}%
\typeout{** the default language instead.}%
\else
\language=\csname l@#1\endcsname
\fi
#2}}
\providecommand{\BIBdecl}{\relax}
\BIBdecl

\bibitem{voice}
X.~Chen, Z.~Li, S.~Setlur, and W.~Xu, ``Exploring racial and gender disparities in voice biometrics,'' \emph{Scientific Reports}, vol.~12, no.~1, p. 3723, 2022.

\bibitem{TOLOSANA2022108609}
R.~Tolosana, R.~Vera-Rodriguez \emph{et~al.}, ``{SVC-onGoing: Signature verification competition},'' \emph{Pattern Recognition}, vol. 127, 2022.

\bibitem{9335993}
R.~Tolosana, R.~Vera-Rodriguez, J.~Fierrez, and J.~Ortega-Garcia, ``{DeepSign: Deep On-Line Signature Verification},'' \emph{IEEE Transactions on Biometrics, Behavior, and Identity Science}, vol.~3, no.~2, pp. 229--239, 2021.

\bibitem{PAMI_RVera2018}
O.~C. Reyes, R.~Vera-Rodriguez, P.~Scully, and K.~B. Ozanyan, ``Analysis of spatio-temporal representations for robust footstep recognition with deep residual neural networks,'' \emph{IEEE Transactions on Pattern Analysis and Machine Intelligence}, no.~99, 2018.

\bibitem{gait1}
P.~Delgado-Santos, R.~Tolosana, R.~Guest, R.~Vera-Rodriguez, and J.~Fierrez, ``{M-GaitFormer: Mobile biometric gait verification using Transformers},'' \emph{Engineering Applications of Artificial Intelligence}, vol. 125, p. 106682, 2023.

\bibitem{gait2}
P.~Delgado-Santos, R.~Tolosana, R.~Guest, F.~Deravi, and R.~Vera-Rodriguez, ``Exploring transformers for behavioural biometrics: A case study in gait recognition,'' \emph{Pattern Recognition}, vol. 143, p. 109798, 2023.

\bibitem{stragapede2022ijcb}
G.~Stragapede, R.~Vera-Rodriguez, R.~Tolosana, A.~Morales, J.~Fierrez, J.~Ortega-Garcia, S.~Rasnayaka, S.~Seneviratne, V.~Dissanayake, J.~Liebers \emph{et~al.}, ``{IJCB 2022} mobile behavioral biometrics competition {(MobileB2C)},'' in \emph{Proc. IEEE International Joint Conference on Biometrics (IJCB)}, 2022.

\bibitem{stragapede2023behavepassdb}
G.~Stragapede, R.~Vera-Rodriguez, R.~Tolosana, and A.~Morales, ``{BehavePassDB: public database for mobile behavioral biometrics and benchmark evaluation},'' \emph{Pattern Recognition}, vol. 134, p. 109089, 2023.

\bibitem{hernandez2019edbb}
J.~Hernandez-Ortega, R.~Daza, A.~Morales, J.~Fierrez, and J.~Ortega-Garcia, ``{edBB: Biometrics and behavior for assessing remote education},'' in \emph{Proc. AAAI Workshop on Artificial Intelligence for Education}, 2019.

\bibitem{rasnayaka2018wants}
S.~Rasnayaka and T.~Sim, ``Who wants continuous authentication on mobile devices?'' in \emph{Int. Conf. on Biometrics Theory, Applications and Systems (BTAS)}, 2018, pp. 1--9.

\bibitem{typenet}
A.~Acien, A.~Morales, J.~V. Monaco, R.~Vera-Rodriguez, and J.~Fierrez, ``{TypeNet: Deep Learning Keystroke Biometrics},'' \emph{IEEE Transactions on Biometrics, Behavior, and Identity Science}, vol.~4, no.~1, pp. 57--70, 2022.

\bibitem{kboc}
A.~Morales, J.~Fierrez, R.~Tolosana, J.~Ortega-Garcia, J.~Galbally, M.~Gomez-Barrero, A.~Anjos, and S.~Marcel, ``Keystroke biometrics ongoing competition,'' \emph{IEEE Access}, vol.~4, pp. 7736--7746, 2016.

\bibitem{Dhakal2018}
V.~Dhakal, A.~M. Feit, P.~O. Kristensson, and A.~Oulasvirta, ``Observations on typing from 136 million keystrokes,'' in \emph{Proc. CHI Conf. on Human Factors in Computing Systems}, 2018.

\bibitem{palin2019people}
K.~Palin, A.~M. Feit, S.~Kim, P.~O. Kristensson, and A.~Oulasvirta, ``How do people type on mobile devices? {Observations} from a study with 37,000 volunteers,'' in \emph{Proc. Int. Conf. on Human-Computer Interaction with Mobile Devices and Services}, 2019.

\bibitem{stragapede2023kvc}
G.~Stragapede, R.~Vera-Rodriguez, R.~Tolosana, A.~Morales, N.~Damer, J.~Fierrez, and J.~Ortega-Garcia, ``{Keystroke Verification Challenge (KVC): Biometric and Fairness Benchmark Evaluation},'' \emph{arXiv:2311.06000}, 2023.

\bibitem{Fierrez-Aguilar2005_ScoreNormalization}
J.~Fierrez-Aguilar, J.~Ortega-Garcia, and J.~Gonzalez-Rodriguez, ``Target dependent score normalization techniques and their application to signature verification,'' \emph{IEEE Trans. on Systems, Man \& Cybernetics - Part C}, vol.~35, no.~3, pp. 418--425, August 2005.

\bibitem{Fierrez-Aguilar2005_AdaptedMultimodal}
J.~Fierrez-Aguilar, D.~Garcia-Romero, J.~Ortega-Garcia, and J.~Gonzalez-Rodriguez, ``Adapted user-dependent multimodal biometric authentication exploiting general information,'' \emph{Pattern Recognition Letters}, vol.~26, no.~16, pp. 2628--2639, December 2005.

\bibitem{ISObiometrics}
\emph{{ISO/IEC 2382-37(en): Information technology — Vocabulary}}, Int. Org. Stand., Geneva, Switzerland, 2022, part 37: Biometrics.

\bibitem{ISO1998}
\emph{{ISO} 9241-11: {E}rgonomic {R}equirements for {O}ffice {W}ork {W}ith {V}isual {D}isplay {T}erminals ({VDT}s)}, Int. Org. Stand., Geneva, Switzerland, 1998, part 11: Guidance on Usability.

\bibitem{gonzalez2021shape}
N.~Gonz{\'a}lez, E.~P. Calot, J.~S. Ierache, and W.~Hasperu{\'e}, ``On the shape of timings distributions in free-text keystroke dynamics profiles,'' \emph{Heliyon}, vol.~7, no.~11, 2021.

\bibitem{gonzalez2022towards}
{N. Gonz{\'a}lez, E. P. Calot, J. S. Ierache, W. and Hasperu{\'e}}, ``Towards liveness detection in keystroke dynamics: Revealing synthetic forgeries,'' \emph{Systems and Soft Computing}, vol.~4, p. 200037, 2022.

\bibitem{2022_PR_SetMargin_Morales}
A.~Morales, J.~Fierrez, A.~Acien, R.~Tolosana, and I.~Serna, ``Setmargin loss applied to deep keystroke biometrics with circle packing interpretation,'' \emph{Pattern Recognition}, vol. 122, p. 108283, February 2022.

\bibitem{gonzalez2023ksdsld}
N.~Gonz{\'a}lez, ``Ksdsld—a tool for keystroke dynamics synthesis \& liveness detection,'' \emph{Software Impacts}, vol.~15, p. 100454, 2023.

\bibitem{vaswani2017attention}
A.~Vaswani, N.~Shazeer, N.~Parmar, J.~Uszkoreit, L.~Jones, A.~N. Gomez, {\L}.~Kaiser, and I.~Polosukhin, ``Attention is all you need,'' \emph{Advances in neural information processing systems}, vol.~30, 2017.

\bibitem{loshchilov2017decoupled}
I.~Loshchilov and F.~Hutter, ``Decoupled weight decay regularization,'' \emph{arXiv:1711.05101}, 2017.

\bibitem{loshchilov2016sgdr}
{I. Loshchilov and F. Hutter}, ``{SGDR: Stochastic gradient descent with warm restarts},'' \emph{arXiv:1608.03983}, 2016.

\bibitem{deng2019arcface}
J.~Deng, J.~Guo, N.~Xue, and S.~Zafeiriou, ``{Arcface: Additive angular margin loss for deep face recognition},'' in \emph{Proc. of the Conf. on computer vision and pattern recognition}, 2019, pp. 4690--4699.

\bibitem{typeformer}
G.~Stragapede, P.~Delgado-Santos, R.~Tolosana, R.~Vera-Rodriguez, R.~Guest, and A.~Morales, ``{TypeFormer}: Transformers for mobile keystroke biometrics,'' \emph{arXiv:2212.13075}, 2023.

\bibitem{FG}
{G. Stragapede, P. Delgado-Santos, R. Tolosana, R. Vera-Rodriguez, R. Guest, and A. Morales}, ``Mobile keystroke biometrics using transformers,'' in \emph{2023 Proc. of Int. Conf. on Automatic Face and Gesture Recognition}, 2023, pp. 1--6.

\bibitem{senerath2023behaveformer}
D.~Senerath, S.~Tharinda, M.~Vishwajith, S.~Rasnayaka, S.~Wickramanayake, and D.~Meedeniya, ``{BehaveFormer: A Framework with Spatio-Temporal Dual Attention Transformers for IMU enhanced Keystroke Dynamics},'' \emph{arXiv:2307.11000}, 2023.

\bibitem{chung2014empirical}
J.~Chung, C.~Gulcehre, K.~Cho, and Y.~Bengio, ``Empirical evaluation of gated recurrent neural networks on sequence modeling,'' \emph{arXiv:1412.3555}, 2014.

\bibitem{watanabe2023tree}
S.~Watanabe, ``Tree-structured parzen estimator: Understanding its algorithm components and their roles for better empirical performance,'' \emph{arXiv:2304.11127}, 2023.

\bibitem{yang2014multimodal}
Q.~Yang, G.~Peng, D.~T. Nguyen, X.~Qi, G.~Zhou, Z.~Sitov{\'a}, P.~Gasti, and K.~S. Balagani, ``A multimodal data set for evaluating continuous authentication performance in smartphones,'' in \emph{Proc. of the ACM Conf. on Embedded Network Sensor Systems}, 2014, pp. 358--359.

\bibitem{acien2021becaptcha}
A.~Acien, A.~Morales, J.~Fierrez, R.~Vera-Rodriguez, and O.~Delgado-Mohatar, ``{BeCAPTCHA: Behavioral bot detection using touchscreen and mobile sensors benchmarked on HuMIdb},'' \emph{Engineering Applications of Artificial Intelligence}, vol.~98, p. 104058, 2021.

\bibitem{userauth}
F.~Bergadano, D.~Gunetti, and C.~Picardi, ``User authentication through keystroke dynamics,'' \emph{ACM Trans. Inf. Syst. Secur.}, vol.~5, no.~4, p. 367–397, 2002.

\bibitem{gonzalez}
N.~Gonzalez, E.~P. Calot, and J.~S. Ierache, ``A replication of two free text keystroke dynamics experiments under harsher conditions,'' in \emph{Int. Conf. of the Biometrics Special Interest Group (BIOSIG)}, 2016, pp. 1--6.

\end{thebibliography}

\end{document}